\documentclass[letterpaper, 10 pt, conference]{ieeeconf}  
\pdfoutput=1

\IEEEoverridecommandlockouts                              

\usepackage{algorithmic}
\usepackage{cite}
\usepackage{graphicx} 
\usepackage{subcaption}
\usepackage{url}
\usepackage{bm}
\usepackage{mathptmx} 
\usepackage{times} 
\usepackage{amsmath} 
\usepackage{amssymb}  
\usepackage{grffile}

\title{\LARGE \bf Efficient Learning of Inverse Dynamics Models \\ for Adaptive Computed Torque Control*
}

\author{David Jorge$^{1}$, Gabriella Pizzuto$^{2}$ and Michael Mistry$^{1}$
\thanks{*This work has received funding from the EPSRC UK RAI Hub National Centre for Nuclear Robotics (NCNR) under agreement EPR02572X/1.}
\thanks{$^{1}$David Jorge and Michael Mistry are with the Institute of Perception, Action and Behaviour
University of Edinburgh, United Kingdom
        {\tt\small s1712653@sms.ed.ac.uk, mmistry@ed.ac.uk}}%
\thanks{$^{2}$Gabriella Pizzuto is with the Department of Chemistry, University of Liverpool, United Kingdom
        {\tt\small gabriella.pizzuto@liverpool.ac.uk}}%
}

\begin{document}

\maketitle
\thispagestyle{empty}
\pagestyle{empty}

\begin{abstract}
Modelling robot dynamics accurately is essential for control, motion optimisation and safe human-robot collaboration. 
Given the complexity of modern robotic systems, dynamics modelling remains non-trivial, mostly in the presence of compliant actuators, mechanical inaccuracies, friction and sensor noise. 
Recent efforts have focused on utilising data-driven methods such as Gaussian processes and neural networks to overcome these challenges, as they are capable of capturing these dynamics without requiring extensive knowledge beforehand.
While Gaussian processes have shown to be an effective method for learning robotic dynamics with the ability to also represent the uncertainty in the learned model through its variance, they come at a cost of cubic time complexity rather than linear, as is the case for deep neural networks.
In this work, we leverage the use of deep kernel models, which combine the computational efficiency of deep learning with the non-parametric flexibility of kernel methods (Gaussian processes), with the overarching goal of realising an accurate probabilistic framework for uncertainty quantification.
Through using the predicted variance, we adapt the feedback gains as more accurate models are learned, leading to low-gain control without compromising tracking accuracy. 
Using simulated and real data recorded from a seven degree-of-freedom robotic manipulator, we illustrate how using stochastic variational inference with deep kernel models increases compliance in the computed torque controller, and retains tracking accuracy.
We empirically show how our model outperforms current state-of-the-art methods with prediction uncertainty for online inverse dynamics model learning, and solidify its adaptation and generalisation capabilities across different setups.

\end{abstract}

\section{Introduction}
\label{section:introduction}
Robotic platforms rely on dynamics models for manipulating objects, motion planning and control. 
Precisely, having a good dynamics model leads to executing trajectories in an accurate yet compliant manner.
Traditionally, the inverse dynamics model was computed using rigid body dynamics equations. 
However, modern systems rely on a vast array of sensors which are noisy, leading to inaccurate models when computed analytically. 
Additionally, unmodelled effects and dynamics such as manufacturing uncertainties, wear and tear, and friction lead to model inaccuracies which could potentially result in undesirable behaviour. 
Non-parametric models address this by adopting a data-driven approach where the model can learn any non-linear function encompassing the forces present in the robot dynamics. 
Here, the problem is viewed as a mapping function from the input states to the output torque.

There exist different regression methods that have been used successfully to learn inverse dynamics models, such as Gaussian processes (GPs)~\cite{4650850} and neural networks (NNs)~\cite{rueckert2017}. 
Gaussian processes offer an accurate prediction with uncertainty estimation represented through the covariance function. 
However, their computational complexity scales exponentially in $O(n^3)$, where $n$ is the set of training data points, and they suffer from quadratic space complexity.
To mitigate this challenge, methods such as local GPR~\cite{4650850}, sparse GP regression (GPR)~\cite{de2012online},~\cite{nguyen2009sparse} and drifting GPs~\cite{7487143} have been introduced. 

Nonetheless, these methods still scale quadratically with the samples ($O(n^2)$) and are thus limited to a few thousand samples. 
A more scalable method is the sparse variational Gaussian process regression (SVGPR)~\cite{hensman2015} method which uses inducing points as an approximation of the full training dataset, with the addition of variational inference to give a tractable objective function.
This achieves good performance on online inverse dynamics learning tasks.


Neural networks excel in approximating non-linear functions and are inherently the most efficient as they come with a time complexity of $O(n)$. 
However, the vanilla deep models lack the existence of a variance for representing the model's prediction uncertainty.
There have been recent efforts to address this, through for example Bayes by Backprop~\cite{blundell2015weight}, which learns a probability distribution over the parameters of a neural network, by minimising the evidence lower bound (ELBO) on the marginal likelihood to optimise the neural network weights.

These variance estimates are essential when learning inverse dynamics models as they improve overall movement representation and are important for adjusting the controller stiffness.
The latter is especially crucial for human-robot collaboration as for safety purposes, robot compliance is key to avoid high motor torques.
This can be achieved through using variable gains that are adapted to the
uncertainty in the model, where accurate tracking is still maintained as the robot is compliant solely whenever the accuracy of the model allows it~\cite{Alberto2014}.

In this work, we propose the sparse variational deep kernel learning (SVDKL) method to achieve this, which combines a highly scalable Gaussian Process framework that is probabilistic in nature, with the adaptable power of deep neural networks.
Specifically, we explore the use of the variance from the SVDKL model for achieving variable-gain control that is computationally efficient for real-time systems.

In summary, the main contributions of this work are as follows: (1) a novel computed torque control model relying on a sparse variational deep kernel learning model for learning efficiently the inverse dynamics models while maintaining compliant control and tracking accuracy, and (2) the evaluation and demonstration of the proposed model for its generalisation ability to different dynamics and across the workspace.







\section{Methodology}
\label{section:methodology}

\subsection{Problem Definition}
\label{subsec:problem_definition}

Model-based robot control utilises forward or inverse dynamics models, which allows for greater tracking performance during tasks, less energy consumption and increased compliance. 
The dynamics of a rigid body in free motion can be empirically formulated through Newton's second law of motion, where for an open-chain robot, its second-order differential equation is:  

\begin{equation}
    \bm{\tau} = \mathbf{M}(\mathbf{q})\mathbf{\ddot{q}}+\mathbf{h}(\mathbf{q},\mathbf{\dot{q}})~.
\label{inversedynamicsequation}
\end{equation}

Equation~\ref{inversedynamicsequation} formulates the robot's inverse dynamics problem which determines the joint forces and torques necessary to attain the robot's desired joint accelerations for a given state. 
$\mathbf{q} \in \mathbb{R}^{n}$ is the vector of joint variables for $n$ degrees of freedom,  $\bm{\tau} \in \mathbb{R}^{n}$ is the vector of joint forces and torques, $\mathbf{M}(\mathbf{q})\in \mathbb{R}^{n \times n}$ is the symmetric positive-definite inertia matrix, $ \mathbf{h}(\mathbf{q}, \mathbf{\dot q})\in \mathbb{R}^{n}$ are forces stemming from the centripetal, Coriolis, gravitational and friction effects.

While this approach is useful for computing the inverse dynamics model, it relies heavily on the underlying model describing the actual system with a high degree of accuracy. 
In practice, this is difficult to obtain as it requires modelling nonlinear physical properties, e.g friction, and any change to the overall environment can cause the underlying physical model to no longer be accurate.
More recent approaches rely on data-driven learning of the robot manipulator dynamics in order to solve Equation~\ref{inversedynamicsequation}. 
Such approaches obtain models that make predictions to track the observed data as closely as possible without assuming any underlying physical structure of the actual system. 
Specifically, an inverse dynamics model is learned by fitting a function $f (\cdot )$, which maps the joint states ($\mathbf{q}$, $\mathbf{\dot q}$, ${\mathbf{\ddot q}}$) to the commanding torques and applying forces $\bm{\tau}$.
For a robotic arm with 7 degrees-of-freedom, this results in a 21-dimensional input and a 7-dimensional output.
This approach is attractive as it allows for any nonlinearities to be compensated for as they are assumed to be inherent in the data~\cite{de2012online}. 


\subsection{Using Deep Kernels for Learning Inverse Dynamics Models}
\label{subsec:methodology_for_dkl}

Instead of relying on the variations of GP or NN models to tackle the inverse dynamics problem, combining both paradigms is key to take advantage of the efficiency of deep neural networks with the non-parametric nature of GP kernel functions. 

To merge both for learning an inverse dynamics model, the Deep Kernel Learning~\cite{wilson2016deep} model can be used, which alters the GP kernel function, $k(\textbf{x}_i, \textbf{x}_j|\theta)$, (where $\textbf{x}_i$ and $\textbf{x}_j$ are the inputs to be transformed, and $\theta$ are the hyperparameters) to:

\begin{equation}
    k(g(\textbf{x}_i,\textbf{w}),g(\textbf{x}_j,\textbf{w})|\mathbf{\theta},\textbf{w}).
    \label{kernel}
\end{equation}

The contribution of the neural network comes from the nonlinear transformation; $g(\textbf{x},\textbf{w})$ is applied to the kernel function arguments as seen in Equation~\ref{kernel}. 
This transformation is learned using a fully connected neural network.

The choice of the kernel function, $k$, depends on the application and problem scope; however a common choice is the RBF kernel~\cite{mcintire2016sparse}, which is smooth and easily tuned, and this is given by:

\begin{equation}
    k(\textbf{x},\textbf{x'})=\sigma^2_f\exp(-\frac{1}{2}\frac{||\textbf{x}-\textbf{x'}||}{\ell^2_d}) ~.
    \label{rbf}
\end{equation}

To learn from an input dataset, $\{\textbf{x}_i,\textbf{y}_i\}^n_{i}$, the augmented kernel function in Equation~\ref{kernel} can be incorporated as the kernel function (covariance function) of any GP~\cite{wilson2016deep}. 
An alternative interpretation is that the neural network acts as a feature extractor, from which a covariance function, $k(X,X')$, from a GP framework is applied to its output layer.

To account for the cubic time complexity of standard GPs, the DKL model makes use of an approximate kernel function with fast computational complexity called the kernel interpolation for scalable structured (KISS)-GP kernel~\cite{wilson2015kernel}, given by:

\begin{equation}
    K_{KISS}=MK_{U,U}M^\top \approx K~,
    \label{KISS}
\end{equation}

which is a separate kernel function that makes use of a sparse matrix, $M$, with interpolation weights, projected on top of the Gaussian Process covariance function, $K$.
This makes online learning possible as training of the KISS-GP scales with $O(n+h(m))$, where $h(m)$ is average with respect to the inducing points, $m$, on average. 

In order to further decrease the time complexity of the DKL model, as well as aiming to achieve a greater and more efficient performance, we make use of the sparse variational Gaussian process framework~\cite{hensman2015}, which is a high performing and scalable model, and combine it with the fully connected neural network feature extractor seen in Equation~\ref{kernel}, to obtain our SVDKL model.

The SVGP framework itself makes use of a set of inducing points, $I_p$, that optimally represent the training data. 
Additionally, variational inference is used to obtain the objective function used to optimise $I_p$ as well as any model hyperparameters~\cite{hensman2015}. 
The objective function is defined as the ELBO on the log marginal likelihood, $\log(\mathbf{y})$.

In order to extend the SVDKL model to a multitask problem, the linear model of coregionalisation (LMC) will be used, which is a general purpose model that assumes that each task, $y_{task}$ is a linear combination of, $D$, latent functions $[g_1, g_2, g_3 ... g_D]$, given by:

\begin{equation}
    y_{task}(\mathbf{x})=\sum_{i=1}^{D} w_i~g_i(\mathbf{x}) ~,
\end{equation}

where $w_i$ are hyperparameters. While LMC allows for dependency between the output tasks, for inverse dynamics learning, the output tasks were made independent.


\subsection{Variable-gain Control}
\label{subsec:vg_control}

In a real-world control system, the feedforward torque would not compensate for unknown system dynamics, and as a result a PD controller would be used to retain tracking accuracy.
The torque output of a PD controller combined with a feedforward model can be defined as:

\begin{equation}
    \bm{\tau} = \mathbf{K}_P(\mathbf{q}_d-\mathbf{q}) + \mathbf{K}_D(\mathbf{\dot{q}_d}-\mathbf{\dot{q}})+\bm{\tau}_{ff} ~,
    \label{control}
\end{equation}

where $\bm{\tau}_{ff}$ is the feedforward torque (given by Equation~\ref{inversedynamicsequation}) and ${\mathbf{K}}_P$ and $\mathbf{K}_D$ are the proportional and derivative feedback controller gains respectively.
In practice, the feedback gains should be increased when the system undergoes unpredictable disturbances. 
Ideally, this should be achieved only where it is necessary to correct the tracking error, as having high feedback gains would affect the system's safety, especially in the presence of humans. 
Thus, accurate models of the feedforward torque allows for compliant low-gain control to be achieved.
 

Alberto et. al~\cite{Alberto2014} exploit the confidence measure of Gaussian Process models in an online inverse dynamics learning setting to control the feedback gains of a PD controller.
This is based upon the notion that having a low model confidence (and thus a high variance) indicates that the predicted torque does not converge to the groundtruth, thus the feedback gains should intuitively be increased in order to compensate for the low confidence of the feedforward model and simultaneously, high model confidence implies that a good model has been learned where low-gain control could be implemented.

Variable-gain control can be achieved with models that have a confidence measure that would allow controlling the contribution of the PD controller torque terms based on the confidence of the model at any timestep. 
This can be further extended by using separate feedback gains for each robot joint such that compliance can act independently for each joint and as a result utilise the independent confidence measure of multi-task GP frameworks. 
In turn, this will consequently lead to an increased position tracking accuracy, as the PD controller will dynamically adapt as the variance changes.

The standard deviation of the model output torque at time $t$, $\sigma(\tau)_t$ is bounded by the model signal variance, $\sigma_f$, and model noise variance, $\sigma_n$.
As such, the feedback torque should be at a maximum when $\sigma(\tau)_t=\sigma_f$ and at a minimum when $\sigma(\tau)_t=\sigma_n$~\cite{Alberto2014}. 
This relationship can be defined as:

\begin{equation}
    K_{P,~t}=K_{min} + (1-z)\cdot(K_{max}-K_{min})
    \label{TSI-KP}
\end{equation}

\begin{equation}
    \textrm{where} \quad z=\textrm{exp}(-C\frac{\sigma(\tau)_t-\sigma_n}{\sigma_f-\sigma_n}).
    \label{TSI-Z}
\end{equation}

$K_{min}$ and $K_{max}$ can be set independently for each joint, by taking into account the torque limits of each joint of the robot manipulator. 
This will objectively keep output torque magnitudes at a controlled level, leading to safe human-robot interaction.
$K_D$ is then related to $K_P$ using $K_D=\zeta\sqrt{K_P}$, where $\zeta$ is the damping ratio.
\section{Experimental Evaluation}
\label{sec:experimental_evaluation}

We evaluated the performance of our model for achieving compliant control and having computational time that allows for the generation of online predictions, without compromising on tracking accuracy.
We posed three questions: (1) how suitable is our model for learning the inverse dynamics model on simulated and real robotic manipulators? (2) how does the model variance change with an increasing number of training data? (3) how do the models generalise?

For all experiments, we used simulated or real data from the KUKA LBR iiwa robotic manipulator.
For the simulation environment, the PyBullet physics engine~\cite{coumans2019} was used due to its open-source license and development process.
For the benchmark datasets, we used $KUKA\_SIM$~\cite{meier2014} and $KUKA\_REAL$~\cite{rueckert2017}.
The first dataset~\cite{meier2014} consists of robotic trajectories recorded in a simulated environment, following rhythmic motions at different speeds.
In the second dataset~\cite{rueckert2017}, the task of the robot was to push a flask that was filled with liquid to a desired goal location, by following a curved trajectory .

We studied four models: our SVDKL and DKL models, Bayesian long short-term memory (B-LSTM) network~\cite{blundell2015weight} which is the current state-of-the-art for deep neural networks with variance estimate, and SVGP model~\cite{hensman2015}, which is a computational efficient variation of GPR.
While we initially considered a GP model, given that our end goal is to deploy the architecture for online learning, using such a model would be computationally not feasible for real world deployment; hence, we decided not to include a GP model in the experimental evaluation.
For all cases, the inputs to the model comprise the joint angles ($radians$), joint velocities ($radians/second$) and joint accelerations ($radians/second^2$), and the outputs are the joint torques ($Newton-metre$).

The implementation of the B-LSTM is in PyTorch~\cite{paszke2019}, whereas the DKL, SVGP, and SVDKL are in PyTorch and GPyTorch~\cite{gardner2018}. 
Experiments were conducted on an Ubuntu machine
with Intel Xeon Gold 6252 (2.1GHz) CPU and a Nvidia RTX 2080 Ti GPU.

\subsection{Experiment I: Learning Inverse Dynamics Models}
\label{ssec:experiment_learning_inverse_dyn_models}
We evaluate the ability of the models to learn the inverse dynamics model of a seven degree-of-freedom robotic manipulator, both in simulation and in the real world.


A subset of the $KUKA\_SIM$~\cite{meier2014} dataset was used, consisting of 330,000 training observations, 90,000 validation observations, and 80,000 testing observations.
The full $KUKA\_REAL$~\cite{rueckert2017} dataset was used, consisting of 386,310 training observations, 96,580 validation observations, and 75,490 testing observations.

On hyperparameters, the DKL and SVDKL share 2 hyperparameters with the SVGP pertaining to the Gaussian Process RBF kernel, namely the signal variance, $\sigma_f$, and the lengthscale, $\ell_d$. 
Likewise, the DKL, SVDKL also share 2 hyperparameters with the B-LSTM, namely the number of neurons, number of hidden layers, and activation functions. 
During offline training, we adopted an early stopping policy (with 15 epochs of patience to prevent over-fitting.

\begin{figure}[h]
    \centering
    \includegraphics[clip, width=0.95\columnwidth]{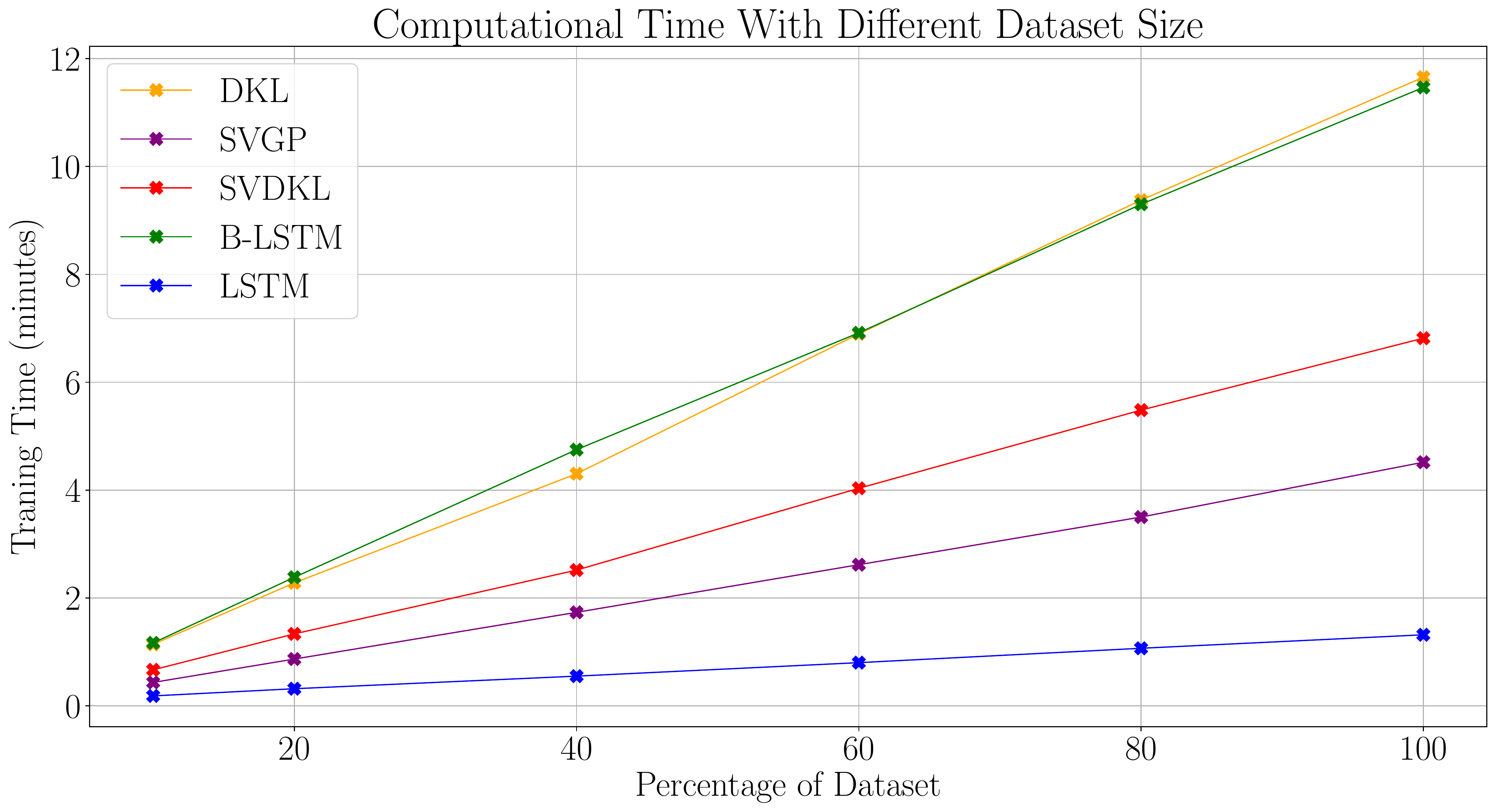}
    \caption{Training time of the DKL, SVDKL, SVGP and B-LSTM models across different sized subsets of the $KUKA\_REAL$~\cite{rueckert2017} dataset.}
    \label{fig:humanoids_train_time}
\end{figure}

The results obtained are depicted in Table~\ref{offline-rmses}, where the average test RMSE for each model is presented. 
The best result is obtained with the B-LSTM, followed closely by the SVDKL and DKL models. 
Across both datasets, the SVGP architecture achieves the worst performance when compared to the other models. 
While the B-LSTM model shows the best performance across both datasets, this comes at a higher computational cost, as illustrated in Fig.~\ref{fig:humanoids_train_time}. 

Specifically, Fig.~\ref{fig:humanoids_train_time} depicts how the developed DKL, SVDKL, SVGP and B-LSTM models scale with an increase in training data. 
While all models have a linear time complexity over the input data,
as evidenced by the results, the SVDKL and SVGP models outperform the DKL and B-LSTM models with respect to training time.
The trade-off obtained by the SVDKL model over the DKL model can also be clearly seen, where while the DKL model is slightly better than the SVDKL in terms of its average test RMSE (Table~\ref{offline-rmses}), the SVDKL model can be trained much faster yet does not compromise significantly on its overall tracking performance.
Thus, the obtained results show empirical evidence that the SVDKL model would not only offer good tracking performance, but can simultaneously be adopted when hundreds of thousands of training observations are available.
The latter is especially important not only for improved scalability, but also crucial for online learning experiments, where the lowest training time is of utmost importance.



\begin{table}[hbt!]
    \centering
    \resizebox{\columnwidth}{!}{
    \begin{tabular}{ccccc}
        \hline
        \textit{Dataset} & \textbf{DKL} & \textbf{SVGP} & \textbf{SVDKL} & \textbf{B-LSTM} \\
        \hline
        \textit{KUKA\_SIM}~\cite{meier2014} & 0.387$\pm$0.0293 & 0.838$\pm$0.0137 & 0.397$\pm$0.0108 & 0.248$\pm$0.0216\\
        \textit{KUKA\_REAL}~\cite{rueckert2017} & 0.829$\pm$0.0005 & 1.412$\pm$0.0011 & 0.852$\pm$0.0007 & 0.654$\pm$0.00018\\
        \hline
    \end{tabular}
    }
    \caption{Average test RMSE and standard deviation for the four models (DKL, SVGP, SVDKL, B-LSTM) averaged over 5 trials.}
    \label{offline-rmses}
\end{table}

\subsection{Experiment II: Understanding the Variance for Compliant Computed Torque Control}
\label{ssec:experiment_vg}

\begin{figure}[hbt!]
\centering
\begin{subfigure}{1.0\columnwidth}
  \centering
  \includegraphics[width=0.9\columnwidth]{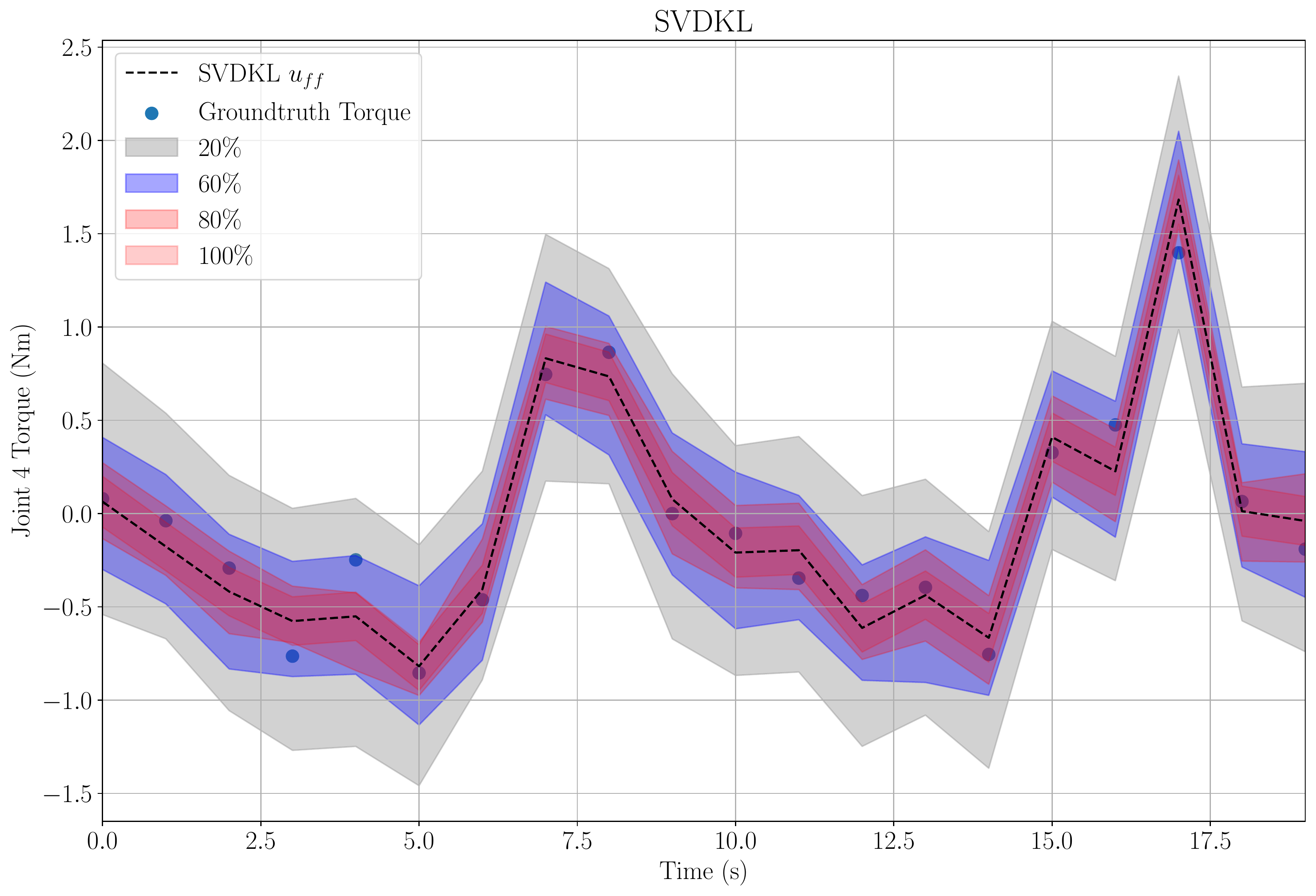}
  \caption{}
  \label{fig:sub1}
\end{subfigure}
\begin{subfigure}{1.0\columnwidth}
  \centering
  \includegraphics[width=0.9\columnwidth]{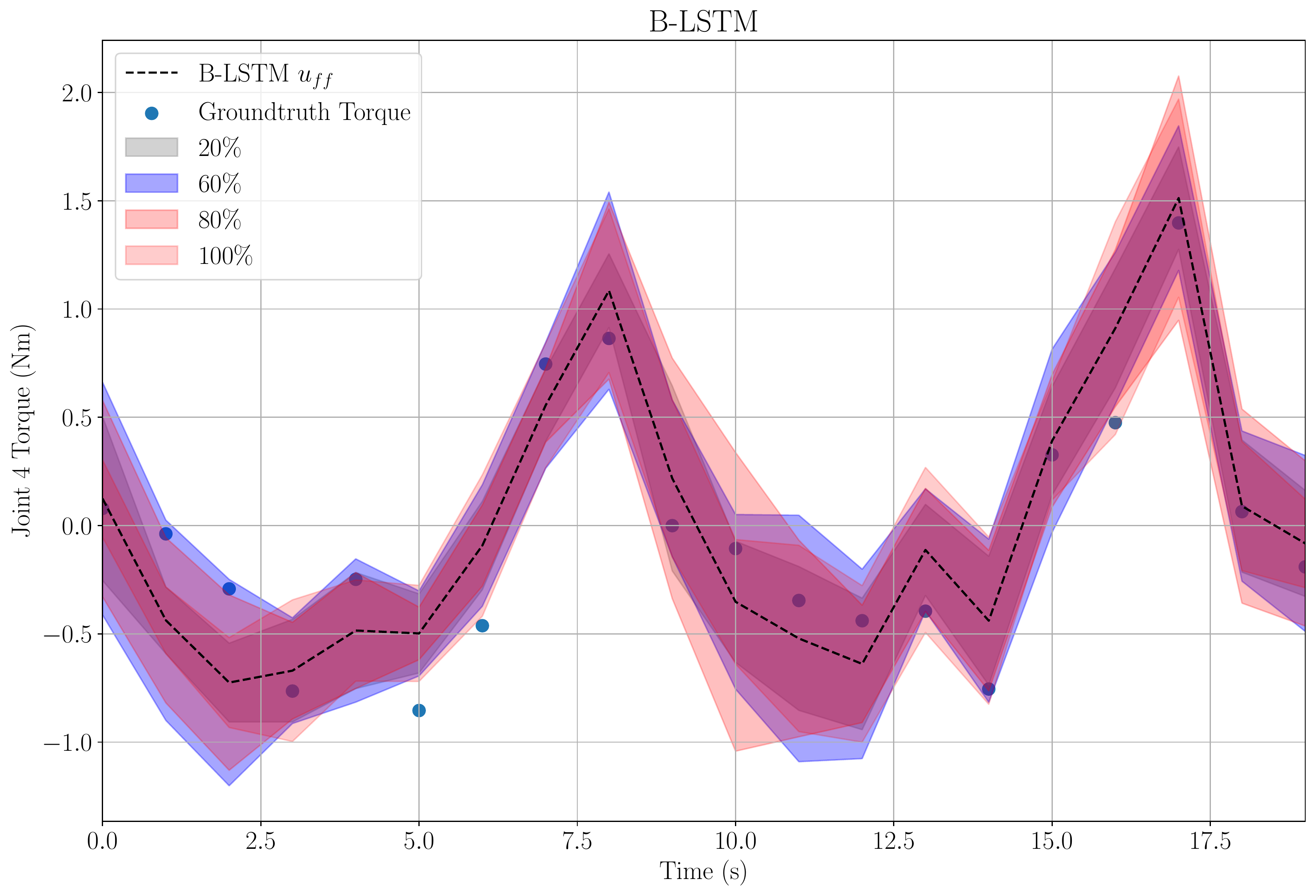}
  \caption{}
  \label{fig:sub2}
\end{subfigure}
\caption{Mean and variance of the SVDKL and B-LSTM models (µ ± 2$\sigma$) on the same test trajectory of 20 seconds, after having been trained with 20\%, 60\%, 80\% and 100\% of the \textit{KUKA\_REAL}~\cite{rueckert2017} dataset. With an increasing number of training data, the variance of the SVDKL is a better measure of confidence compared to the B-LSTM.}
\label{fig:online-joint-rmse}
\end{figure}

As illustrated in related works~\cite{Alberto2014}, model variance could be used as a measure to achieve low-gain control as more accurate models are obtained. 
To date, this has only been realised through using GPR, which is not feasible for online learning.
Alternatively, SVGPR could be used at the expense of significantly lower tracking accuracy.
Here, we proposed the use of deep kernel models to keep the variance from the GP which is also fast enough for real-time requirements on robotic systems.
We compared this to Bayesian neural networks, specifically the B-LSTM, to understand better whether these variance estimates would allow us to achieve low-gain control when an accurate model is obtained.
Our results are depicted in Fig.~\ref{fig:online-joint-rmse}, where we compare the mean and variance of the SVDKL and B-LSTM models with an increasing number of training trajectories. 
As illustrated, the variance for the SVDKL decreases as the model prediction converges towards its true value.
On the otherhand, as the predictive variance for the B-LSTM is solely an estimate, it does not decrease as learning progresses.
Hence, the variance for the latter would not be a good indication for lowering the gains when delegating from the feedback to the feedforward terms.
In fact, we studied how the feedback gains would vary should we use the model variance.
It was evident that the SVDKL model achieves better stability with the variable gain controller. 
As a result, we opted to not pursue the B-LSTM model in further evaluations as it would not be possible to use it for achieving low-gain control as the model error decreases.


\subsection{Experiment III: On Generalisation}
\label{ssec:experiment_generalisation}

\begin{figure}[h]
\centering
\begin{subfigure}{0.5\columnwidth}
  \label{subfig:sub_cube_pickup_setup}
  \centering
  \includegraphics[width=0.99\columnwidth]{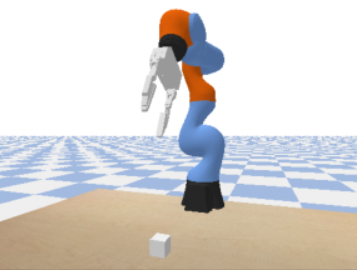}

\end{subfigure}%
\begin{subfigure}{0.5\columnwidth}
  \centering
  \includegraphics[width=0.7\columnwidth]{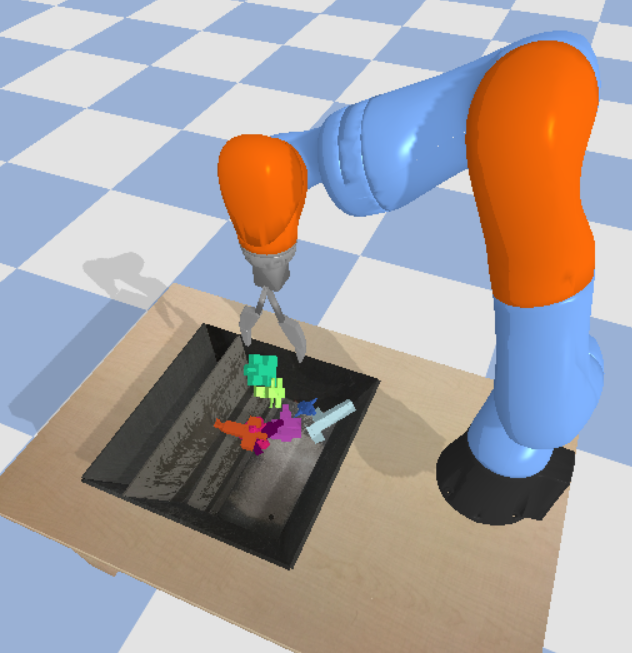}
  \label{subfig:random_objects_pickup_setup}
\end{subfigure}%
\caption{An overview of the simulation setup with the KUKA LBR iiwa robotic manipulator. Different target objects (cube and randomly-shaped blocks) are used to study adaptation to change in dynamics.}
\label{fig:online-setup-environment}
\end{figure}

A major concern of data-driven models is that they are significantly challenged when the test data is outside of the distribution of the training dataset.
To address this, we looked into how our models would generalise under different loads and when the test trajectory is in a different task space from the training trajectories.
Our setup comprised the robotic platform on a table with different objects (cubes and randomly shaped blocks).
In particular, the models were studied in two different setups, across three tasks.
Fig.~\ref{fig:online-setup-environment} illustrates both setups: the pick and place task with a cube, where the robotic manipulator picks up a cube from the centre of the table and drops it on its left-hand side and the setup for pick and place of random objects (generated using the PyBullet random urdf files library), where the KUKA IIWA robot grasps a random object and drops it to its left-hand side. 
The randomly-generated objects were chosen to be irregular to challenge the task dynamics.
The three tasks were the following: (1) pick and place of cube with different mass (1-15 kg); (2) pick and place of randomly-shaped object and (3) pick and place with a `mirrored cube' with different mass (1-15 kg), where the models were first trained using the original pick and place task, and then evaluated on a mirrored version of the task (this experiment was carried out to evaluate how the model generalises to a different task space).

Worth noting is that each setup makes use of different grippers: the setup on the left in Fig.~\ref{fig:online-setup-environment} uses a parallel gripper, while the setup on the right uses a tong gripper. 
By doing so, we demonstrate how the proposed approach would generalise to different setups and experimental conditions. 

For all experiments, each feedforward model was warm-started with training trajectories from~\cite{meier2014}, after which it was trained in batches of size 100 during the course of the experiment. 
Before each training batch, the feedforward model predicts the joint torques, followed by training on data gathered from the combined feedforward and feedback torque and the recorded robot states. 
A PD controller with variable gains was used to correct any position error that could arise in the presence of model imperfections. 

\begin{figure}[hbt!]
    \centering
    \includegraphics[width=0.99\columnwidth]{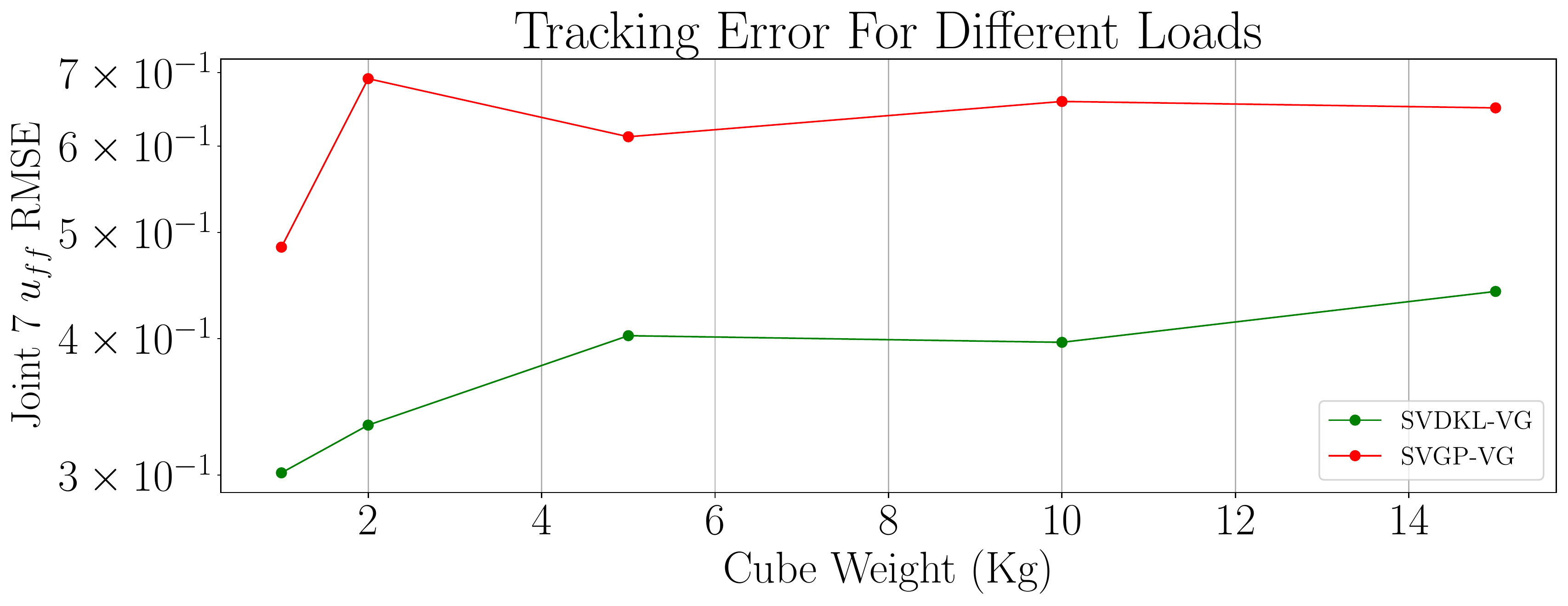}
    \caption{The RMSE for the KUKA LBR iiwa end effector during pick and place of cubes with different weight, using the SVDKL and SVGP models with variable gains.}
    \label{cube-weights}
\end{figure}

\begin{figure}[hbt!]
\centering
\begin{subfigure}{1.0\columnwidth}
  \centering
  \includegraphics[width=0.95\columnwidth]{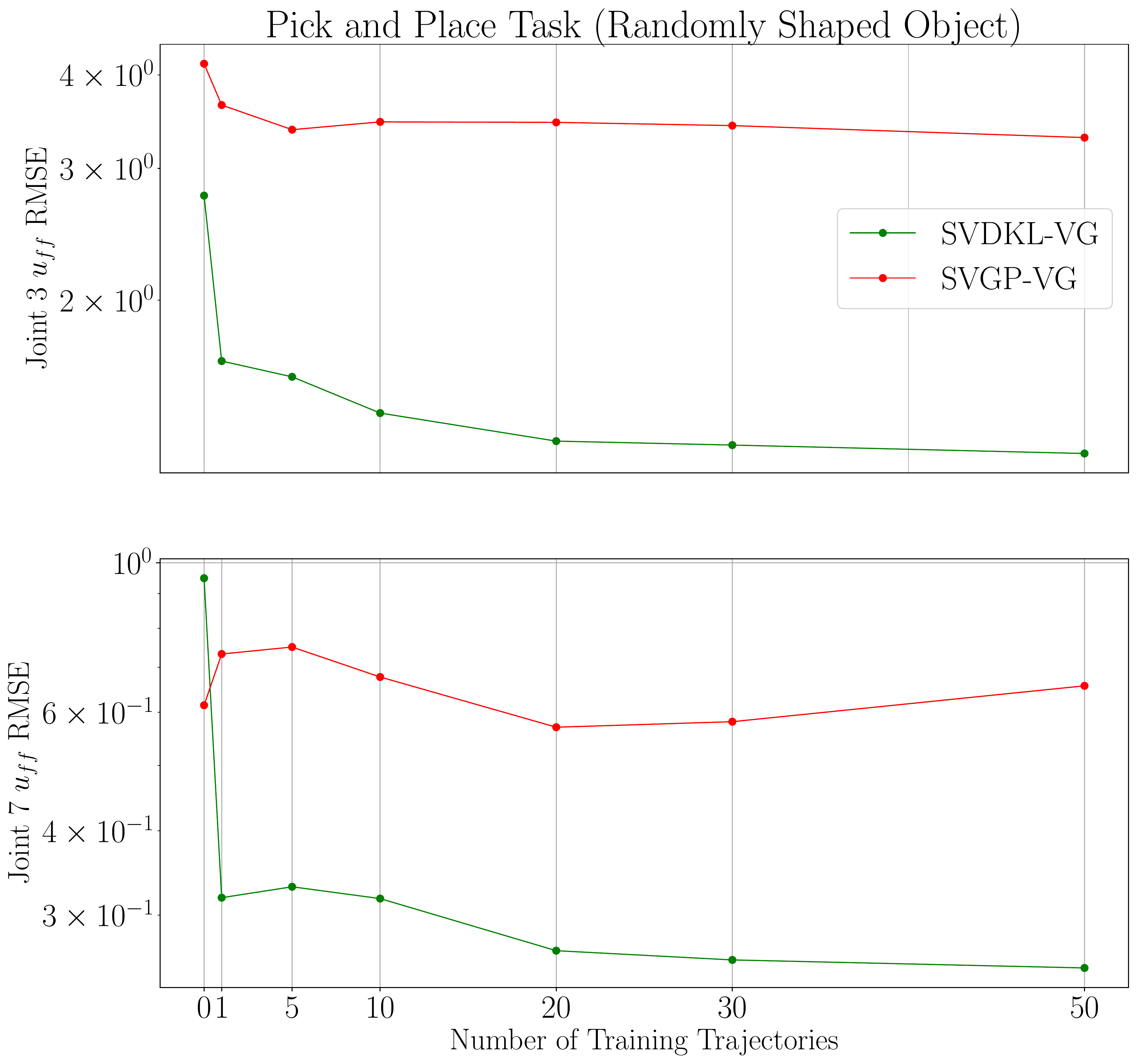}
  \label{fig:sub2}
\end{subfigure}
\caption{RMSE for joints 3 and 7 of the KUKA LBR iiwa robot manipulator on the pick and place task using randomly-shaped objects. Using the same models (SVDKL and SVGP) with variable gains, we observed similar trends across the other two tasks.}
\label{online-joint-rmse}
\end{figure}

Table~\ref{online-rmses} outlines the RMSE of the end effector (joint 7) for each model across the different tasks. 
It can be observed that the SVDKL model outperforms the SVGP model across all tasks, where the mean and variance are both significantly lower.
Fig.~\ref{cube-weights} shows the feedforward torque RMSE of the end effector for different cube weights. 
Overall, the model performances deteriorate as the mass of the cube increases, and hence it is crucial that the feedback gains reflect this change whilst maintaining overall stability.
Nonetheless, the SVDKL model demonstrates lower RMSE for the predicted torque, even with varying loads.
Fig.~\ref{online-joint-rmse} shows how the RMSE for joint 3 and joint 7 (the end effector) change with an increased number of training trajectories for each experiment. 
As more training trajectories are used, the SVDKL model learns the system dynamics more efficiently than the SVGP model. 



\begin{table}[hbt!]
    \centering
    \resizebox{\columnwidth}{!}{
    \begin{tabular}{cccc}
        \hline
        \textit{Task} & \textbf{SVGP} & \textbf{SVDKL} \\
        \hline
        Pick and place (cube) & 0.0170$\pm$0.0254 & 0.00869$\pm$0.00173 \\
        Pick and place (randomly-generated object) & 0.784$\pm$1.296 & 0.366$\pm$0.123 \\
        Pick and place (`mirrored cube') & 0.102$\pm$1.076 & 0.0148$\pm$0.0509 \\        
        \hline
    \end{tabular}
    }
    \caption{RMSE of the KUKA LBR iiwa position for the end effector (joint 7) in each of the experiments averaged over 5 trials.}
    \label{online-rmses}
\end{table}

\subsection{Discussion}
\label{subsec:discussion}

Overall, the experiments carried out do not solely validate how our proposed SVDKL method is capable of learning high-dimensional inverse dynamics models, but also illustrate the benefits of the tandem GP with NN models namely their performance which is comparable to other state-of-the-art methods, runtime, and scalability for real-world deployment.
We believe that our approach demonstrated that computationally efficient methods can not only give good tracking performance, as opposed to the SVGP model, but can also be capable of having accurate variance that would allow the robot to become compliant when the task allows it.
In fact, for the pick and place task of randomly-shaped objects with a warmstart of 50 trajectories, the SVDKL variable-gain controller was able to decrease the feedback gains across all joints by 18.54\%.
We consider this an important step towards robotic systems that are able of autonomously and repeatedly adapting their feedforward and feedback parameters with the overarching goal of having safe physical human-robot interaction.






\section{Conclusion}
\label{section:conclusion}

In this work, we demonstrate how to leverage the strengths of Gaussian processes and neural networks in tandem to achieve similar performance to state-of-the-art methods for learning inverse dynamics models, while simultaneously obtaining a reliable confidence measure in their output.
In fact, our proposed sparse variational deep kernel learning model was demonstrated to have both the prediction error and the variance decrease exponentially with the number of samples.
Moreover, our framework outperformed the state-of-the-art methods for uncertainty estimation in online learning environments -- sparse variational Gaussian processes.
Empirically we show how this approach surpasses variance estimates given by neural networks with respect to the uncertainty of the model.
We evaluate our approach on different experimental setups using a seven degree-of-freedom robotic manipulator, specifically in terms of its ability of adapting to changes in dynamics and generalising to a new task space.
Our experimental evaluation offers compelling evidence that there are benefits to using our method for compliant control, whilst simultaneously decreasing tracking errors. 

As we move closer towards systems that adapt to real-world conditions, our novel framework provides a seemingly simple yet elegant method to achieve task-specific compliance, that would lead to safer human-robot collaboration.
In future endeavours, we will further analyse the overall effect that varying the gains would have on the stability of the system and we plan on exploring how using physics-guided kernels would improve the overall learning process, especially in the presence of contacts~\cite{pizzuto2021}.






\bibliographystyle{ieeetr}
\bibliography{references.bib} 

\end{document}